\pdfoutput=1

\documentclass[11pt]{article}

\usepackage[final]{acl}
\usepackage{multirow}
\usepackage{times}
\usepackage{latexsym}

\usepackage[T1]{fontenc}

\usepackage[utf8]{inputenc}

\usepackage{microtype}

\usepackage{inconsolata}

\usepackage{graphicx}
\usepackage{amsmath}
\usepackage{subfigure}
\usepackage{colortbl}  

\def\method{ORION}

\title{Enhancing Long-Chain Reasoning Distillation through Error-Aware Self-Reflection}

\author{Zhuoyang Wu$^{1}$\thanks{ \ \ indicates equal contribution.}, Xinze Li$^{1}$\footnotemark[1], Zhenghao Liu$^{1}$\thanks{ \ \ indicates corresponding author.}, Yukun Yan$^{2}$,\\ \textbf{Zhiyuan Liu$^{2}$, Minghe Yu$^{1}$, Cheng Yang$^{3}$, Yu Gu$^{1}$, Ge Yu$^{1}$, Maosong Sun$^{2}$} \\ 
$^1$School of Computer Science and Engineering, Northeastern University, China \\
$^2$Department of Computer Science and Technology, Institute for AI, Tsinghua University, China \\
$^3$School of Computer Science, Beijing University of Posts and Telecommunications, China\\
}

\begin{document}
\maketitle

\begin{abstract}
Large Language Models (LLMs) have exhibited strong reasoning capabilities and achieved remarkable performance in mathematical problem-solving tasks.
Recently, distilling reasoning ability from long-form Chains-of-Thought (CoTs) has emerged as a promising approach for enhancing Small Language Models (SLMs). Existing studies typically treat SLMs as student models and use long-form CoTs as supervision signals for Supervised Fine-Tuning (SFT) to transfer reasoning ability.
However, such long-form CoT teachers are usually unaware of the student model's capacity, which limits the effective utilization of the provided reasoning traces.
To overcome this limitation, we propose err\textbf{O}r-aware self-\textbf{R}eflect\textbf{ION} (\method), a framework that refines teacher CoTs through an Error-Aware Reflection process. \method{} enables the student model to construct more tailored teacher CoTs by refining teacher CoTs and incorporating its own reasoning errors.
Experiments on multiple mathematical reasoning benchmarks demonstrate that \method{} consistently improves performance by more than 2\% over all baselines. Further analysis reveals that the CoTs constructed by \method{} exhibit higher coherence and logical consistency, thereby serving as more effective supervision signals for SFT.
All codes are available at \url{https://github.com/NEUIR/ORION.git}.
\end{abstract}


\section{Introduction}
\begin{quote}
    \emph{Success does not consist in never making mistakes, but in never making the same one a second time.}

    \hfill — George Bernard Shaw
\end{quote}

\begin{figure}[t]
    \centering
    \includegraphics[width=1\linewidth]{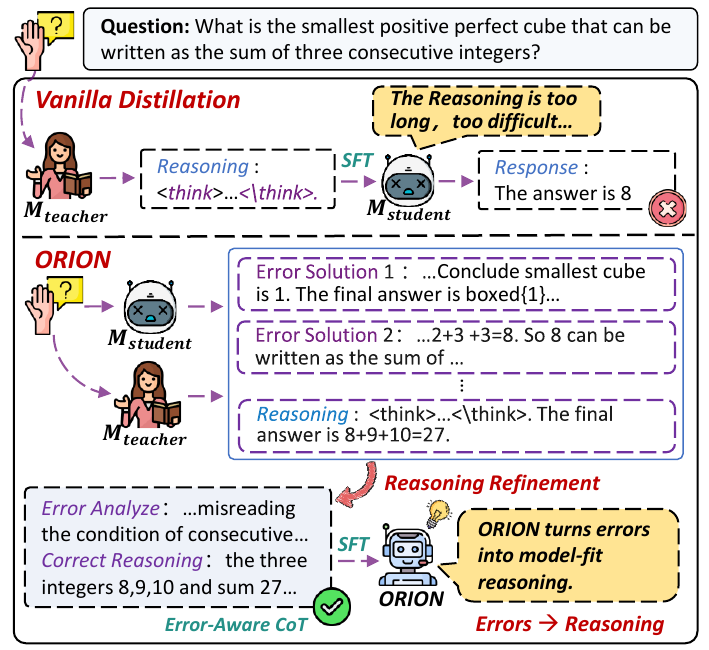}
    \caption{The Framework of Our \method{} Model. \method{} refines long-form reasoning via self-reflection.}
    \label{fig:enter-label}
\end{figure}







\noindent Large Language Models (LLMs) have demonstrated remarkable reasoning capabilities in mathematical problem-solving~\cite{brown2020languagemodelsfewshotlearners, zhang2022optopenpretrainedtransformer}. By leveraging the Chain-of-Thought (CoT) paradigm~\cite{wei2022chain}, the LLMs can decompose and solve mathematical problems step by step~\cite{fu2023specializing,li2023query}, and achieve better performance. However, for Small Language Models (SLMs), generating faithful and logically coherent CoTs remains a significant challenge, particularly when solving the complex mathematical tasks~\cite {wang2025comprehensivesurveytrustworthinessreasoning}.

Recent studies have focused on distilling reasoning abilities from long-form CoTs to enhance the performance of SLMs on mathematical reasoning tasks~\cite{ye2025limoreasoning,deepseekai2025deepseekr1incentivizingreasoningcapability}. These methods typically regard SLMs as student models and employ long-form CoTs generated by Reasoning Language Models (RLMs) as supervision signals for Supervised Fine-Tuning (SFT), thereby enabling the transfer of reasoning capabilities from teacher to student models~\cite{hsieh2023distillingstepbystepoutperforminglarger}. However, during the distillation process, the substantial capability gap between student and teacher models often leads SLMs to passively absorb knowledge from the teacher~\cite{zhou2022bertlearnsteachknowledge}, making it difficult for them to internalize overly complex or lengthy reasoning steps and thereby limiting the effectiveness of knowledge transfer~\cite{li2025smallmodelsstrugglelearn,anonymous2025morale}.

To address this challenge, recent studies have explored error-aware learning, which leverages the solution errors generated by SLMs to train them to avoid similar mistakes, thereby improving their mathematical question answering capabilities~\cite{an2023learning}. These approaches typically employ superior LLMs to revise the incorrect solutions generated by SLMs, treating the corrected solutions as training targets while using the original erroneous solutions as input to teach SLMs to reflect on their mistakes and avoid repeating them during inference~\cite{an2023learning,li2024learning}. However, these methods do not account for the student’s own errors during the CoT refinement process, neglecting crucial signals that could be used to construct more tailored supervision when distilling reasoning capabilities from the CoTs.


In this paper, we introduce err\textbf{O}r-aware self-reflect\textbf{ION} (\method{}), which constructs more effective supervisions to help distill the reasoning capabilities of a teacher model, such as Deepseek-R1~\cite{deepseekai2025deepseekr1incentivizingreasoningcapability}, into a student model implemented as SLMs via SFT. To generate more effective supervision signals, \method{} leverages the student model itself to refine the long-form CoTs produced by LRMs through self-reflection, thereby narrowing the capability gap and enabling the student to better emulate the teacher's reasoning patterns. Moreover, solution errors made by the student are included in the prompts and considered during the refinement process, guiding the student model to produce more adaptive and accurate long-form CoTs for supervision.




Our experiments show that \method{} outperforms all baselines across mathematical tasks of varying difficulty, demonstrating its overall effectiveness. Further analysis indicates that, compared to using long-form CoTs alone as the training target, \method{} achieves lower and more stable variance-entropy scores during training, highlighting its improved training stability. Moreover, \method{} effectively eliminates redundant reasoning patterns in long-form CoTs, enabling SLMs to learn more concise and higher-quality reasoning processes. Finally, we observe that \method{} corrects various types of errors generated by vanilla SLMs, particularly reasoning errors, further validating the effectiveness of its error-aware CoT refinement mechanism.

\section{Related Work}

Large Language Models (LLMs) have proven effective in solving mathematical problems~\cite{cobbe2021trainingverifierssolvemath,hendrycks2021measuringmathematicalproblemsolving}. Advanced prompting strategies, such as Chain-of-Thought (CoT)~\cite{wei2022chain}, decompose the problem-solving process of LLMs into a series of intermediate steps, significantly enhancing their mathematical reasoning abilities~\cite{li2023query,qin2023cross,luo2023wizardmath}. To further enhance the performance of LLMs, particularly for Small Language Models (SLMs), some studies utilize human-annotated solution labels to fine-tune the models~\cite{hsieh2023distillingstepbystepoutperforminglarger}. However, this approach often leads to model overfitting to the training labels, limiting the model's ability to learn genuine reasoning knowledge from supervision signals~\cite{luo2023empirical, gudibande2023false}.

To construct high-quality SFT datasets, recent studies typically explore long-form reasoning distillation methods~\cite{yang2025qwen3technicalreport,ye2025limoreasoning}. Specifically, these methods treat Reasoning Language Models (RLMs), such as Deepseek-R1~\cite{deepseekai2025deepseekr1incentivizingreasoningcapability}, as teacher models and use their generated long-form CoTs as supervision targets to fine-tune SLMs, which enables the student model to learn not only the diverse mathematical solution strategies but also the underlying reasoning patterns~\cite{li2025llms,DBLP:conf/iclr/0006YZXG0Y25}. However, several insightful studies indicate that due to the capacity and capability limitations of SLMs, these models struggle to effectively learn reasoning strategies from the long-form CoTs generated by more capable teacher models~\cite{li2025smallmodelsstrugglelearn,anonymous2025morale}, and may even encounter issues such as content repetition and excessive reflection~\cite{li2025smallmodelsstrugglelearn}.

Different from directly distilling long-form reasoning trajectories from teacher models, another research employs superior LLMs to correct and analyze the error reasoning trajectories generated by SLMs, thereby constructing high-quality SFT datasets~\cite{tong2024llmslearnpreviousmistakes,pan2025lemma}. Specifically, \citet{an2023learning} propose a reasoning error correction SFT method, where incorrect reasoning generated by SLMs is used as input, and the SLMs are tasked with reproducing the correction results generated by a superior LLM. \citet{tong2024llmslearnpreviousmistakes} utilize multiple teacher models to generate guided learning data for correcting the error solutions generated by the student model, thereby enabling the student model to learn both correct answers and avoid errors during SFT. However, these methods only optimize the model to avoid repeating errors and do not incorporate the student model's error solutions during the CoT refinement process, making it difficult to construct high-quality supervision signals for distilling reasoning capabilities. In contrast, \method{} focuses on enhancing reasoning distillation through error-aware self-reflection, enabling SLMs to reflect on the generated errors and refine long-form reasoning data for SFT.

\begin{figure*}[t]
    \centering
      \small\includegraphics[width=\linewidth]{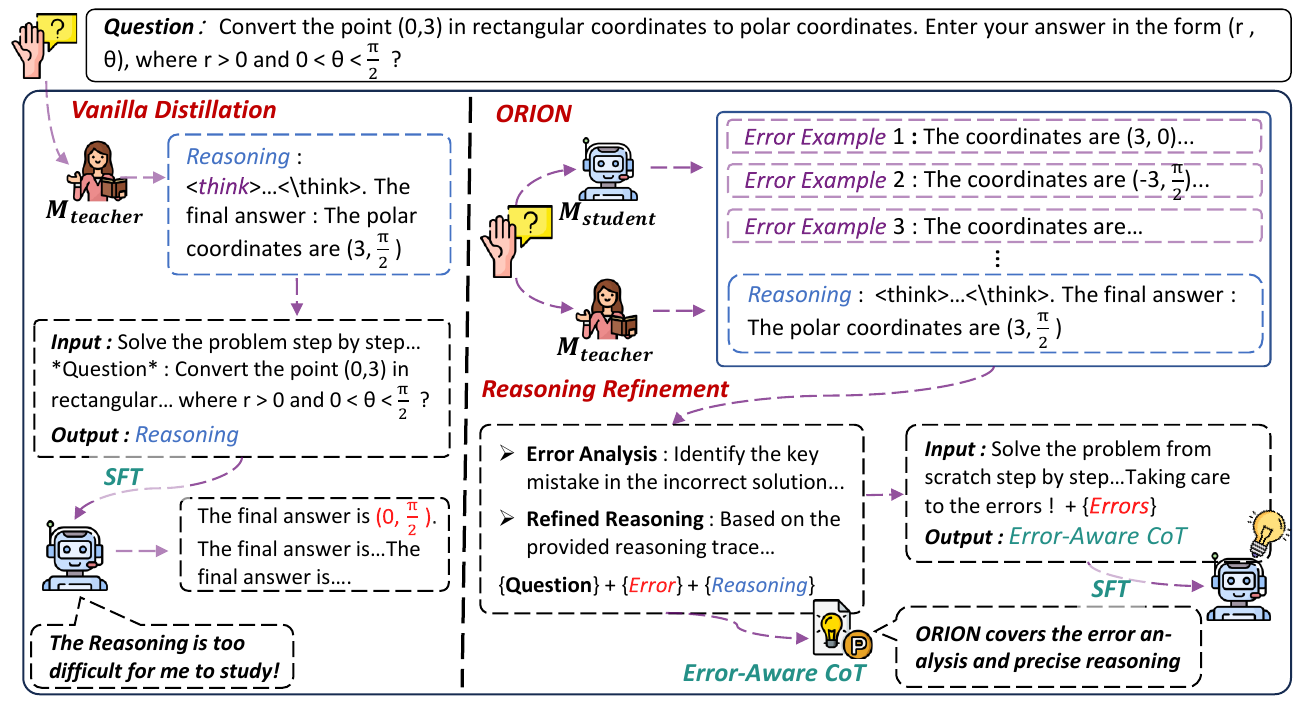}
    \caption{Illustration of Our \method{} Model.}
    \label{fig:model}
\end{figure*}












\section{Methodology}
In this section, we present Error-Aware Self-Reflection (\method) in Figure~\ref{fig:model}, a method designed to enhance the SFT of LLMs for mathematical reasoning.
We first describe how to distill knowledge from long-form chain-of-thought (CoT) traces using standard SFT (Sec.~\ref{sec:3.1}). We then introduce our error-aware data refinement framework (Sec.~\ref{sec:3.2}), which identifies potential reasoning errors and incorporates them into CoT outputs, thereby enhancing the quality of SFT datasets.

\subsection{Distilling Long Reasoning Capability from LLM via Supervised Fine-Tuning}\label{sec:3.1}
Given a mathematical question $q$, the LLM is prompted to generate a solution using a task-specific instruction ($\text{Instruct}_\text{QA}$). To enhance the reasoning capabilities of LLMs, we consider a Small Language Model (SLM) as the student model $\mathcal{M}_s$ and explore two SFT strategies to optimize its parameters $\theta$: vanilla SFT and reasoning distillation via long-form reasoning traces.

\textbf{Vanilla SFT Method.}
In the vanilla SFT paradigm, the model is fine-tuned on human-annotated solution labels $\mathcal{D}_\text{Raw} = \{(q^1, L^1), \dots, (q^n, L^n)\}$~\cite{wang2023making}, which serve as ground-truth outputs for each input question $q^i$. The training objective minimizes the negative log-likelihood of the ground truth solution $L^i$:
\begin{equation}\small
 \mathcal{J} = -\sum_{i=1}^{n} \sum_{t=1}^{|L^i|} P(L_t^i|L_{<t}^i,\text{Instruct}_\text{QA}(q^i);\theta).
\end{equation}
While this approach has demonstrated strong empirical performance in enhancing the problem-solving accuracy of LLMs~\cite{hsieh2023distillingstepbystepoutperforminglarger}, it often overfits to the training labels, thereby limiting the student model's ability to acquire genuine reasoning knowledge from supervision~\cite{luo2023empirical,gudibande2023false}. Consequently, recent research has increasingly focused on curating high-quality datasets specifically designed for SFT~\cite{wettig2024qurating}.

\textbf{Long-form Reasoning Distillation.}
Unlike the standard SFT approach, the reasoning distillation method incorporates Chain-of-Thought (CoT) supervision to guide the model through intermediate reasoning steps, rather than directly optimizing for the final label~\cite{hsieh2023distillingstepbystepoutperforminglarger,deepseekai2025deepseekr1incentivizingreasoningcapability}.
This paradigm encourages LLMs to internalize not only the correct solution but also the underlying logical structure and reasoning patterns demonstrated by the teacher model $\mathcal{M}_t$.

Specifically, given a question $q$, we first regard a Large Reasoning Model $\mathcal{M}_t$, such as DeepSeek-R1~\cite{deepseekai2025deepseekr1incentivizingreasoningcapability}, as the teacher model and prompt it to generate a Chain-of-Thought (CoT) style response $o$:
\begin{equation}\small\label{eq:error}
o = \mathcal{M}_t(\text{Instruct}_\text{QA}(q)).
\end{equation}
We then construct the SFT dataset $\mathcal{D} = {(q^1, o^1), \dots, (q^n, o^n)}$ by pairing each input query $q^i$ with its corresponding CoT-style response $o^i$ generated by the teacher model $\mathcal{M}_t$.
The distilled CoT response $o^i$ is used as the supervision target for fine-tuning the student model $\mathcal{M}_s$, enabling it to acquire step-by-step reasoning abilities from the teacher's intermediate reasoning traces:
\begin{equation}\small\label{eq:longcot-sft}
 \mathcal{J} = -\sum_{i=1}^{n} \sum_{t=1}^{|o^i|}P(o_t^i|o_{<t}^i,\text{Instruct}_\text{QA}(q^i);\theta).
\end{equation}
This distillation approach encourages the student model to internalize the reasoning behaviors of more capable teacher models, thereby fostering the generation of coherent and logically grounded CoT outputs.
However, since the teacher model is unaware of the student’s capacity, directly fine-tuning on long-form reasoning traces can be challenging for student models.
Naively distilling from the teacher may lead to degenerate behaviors such as repetitive outputs or unstable training dynamics~\cite{yin2025marcoo1v2wideningdistillation,li2025smallmodelsstrugglelearn}.
To address these issues, \method{} refines long-form CoTs by error-aware self-reflection, providing more tailored and effective training signals for SFT (Sec.~\ref{sec:3.2}).

\subsection{Refining Long-Form Reasoning via Error-Aware Self-Reflection}\label{sec:3.2}
To enhance the quality of long-form reasoning data for SFT, \method{} introduces an error-aware reflection mechanism that leverages the student model's own mistakes and self-reflection to refine the teacher-provided long-form CoTs.

Given a dataset $\mathcal{D} = \{(q^1, o^1), \dots, (q^n, o^n)\}$, where $q^i$ is a mathematical question and $o^i$ is a corresponding CoT-style solution generated by the teacher model $\mathcal{M}_t$, \method{} first prompts the student model $\mathcal{M}_s$ to generate multiple potentially incorrect solutions $Y^e$ for each $q^i$ (Sec.~\ref{method:error_exposure}).
Then, these erroneous outputs expose the model's failure modes and are used to guide a reflection-based refinement process, where $\mathcal{M}_s$ is prompted to revise the original teacher CoT-style response $o^i$ into a more robust solution $\tilde{o}^i$ by explicitly avoiding previously observed errors (Sec.~\ref{method:self_refine}). The resulting refined dataset $\tilde{\mathcal{D}}$ is then used for SFT, as in Eq.~\ref{eq:longcot-sft}, to further improve the reasoning capability of the student model $\mathcal{M}_s$.

\subsubsection{Error Exposure via Response Sampling}\label{method:error_exposure}
To uncover typical reasoning errors made by the student model $\mathcal{M}_s$, we adopt a response sampling strategy inspired by prior work~\cite{an2023learning}. For each input question $q^i$, we generate a diverse set of $K$ candidate solutions $Y^i = \{y_1^i, \dots, y_K^i\}$ by sampling from $\mathcal{M}_s$ under a range of temperatures $\tau$:
\begin{equation}\small\label{eq:error}
Y^i \sim 
\text{Sample}_ \tau(\mathcal{M}_s(\text{Instruct}_\text{QA}(q^i))).
\end{equation}
Then each sampled response $y_k^i \in Y^i$ is then post-processed using an answer extraction function $\text{Ans}(\cdot)$, and its correctness is checked against the ground truth label $L^i$. These responses that produce incorrect final answers are collected into an error set $Y^i_\text{err}$:
\begin{equation}\small\label{eq:filter}
Y^i_{\text{err}} = \{ y_k^i \mid \text{Ans}(y_k^i) \ne L^i \}.
\end{equation}
The erroneous solutions, denoted as $Y^i_{\text{err}} = \{ y^i_1, \dots, y^i_m \}$, serve as references for subsequent reflection, helping the student model identify failure patterns and revise the teacher-generated CoT traces accordingly.

\subsubsection{Reasoning Refinement through Error-Aware Self-Reflection}\label{method:self_refine}
To help the student model $\mathcal{M}_s$ internalize more robust reasoning heuristics during SFT, we introduce a self-reflection mechanism for the student model $\mathcal{M}_s$, which leverages its own erroneous responses to improve the quality of SFT data.
Specifically, given an initial long reasoning-based SFT dataset $\mathcal{D} = \{(q^1, o^1), \dots, (q^n, o^n)\}$, the student model $\mathcal{M}_s$ is encouraged to analyze its own mistakes and learn to refine the original CoT outputs for SFT.

For each mathematical query $q^i$ in $\mathcal{D}$, we begin with an initial long-form CoT-style response $o^i$ and a set of model-generated erroneous attempts $Y^i_{\text{err}} = \{ y^i_1, \dots, y^i_m \}$. For each error example $y^i_k \in Y^i_{\text{err}}$, the student model $\mathcal{M}_s$ is prompted to refine the response $o^i$ generated by the teacher model $\mathcal{M}_t$ by reflecting on these self-made errors:
\begin{equation}\small\label{eq:error}
\tilde{o}^i_k = \mathcal{M}_s(\text{Instruct}_\text{Ref}(q,y^i_k,o^i)),
\end{equation}
where $\text{Instruct}_\text{Ref}$ denotes a self-reflection instruction that guides the student model $\mathcal{M}_s$ to analyze and refine $o^i$ based on its error response $y^i_k$.
This yields a set of refined reasoning responses for the mathematical query $q^i$:
\begin{equation}\small
    \tilde{O}^i = \{ \tilde{o}^i_1, \dots, \tilde{o}^i_m \}.
\end{equation}
To ensure correctness and eliminate spurious refinements, we filter out any $\tilde{o}^i_k$ whose final answer does not match the ground-truth label $L^i$:
\begin{equation}\small\label{eq:filter}
\tilde{\mathcal{D}}^i = \{ (q^i, \tilde{o}^i_k) \mid 1 \le k \le m,\ \text{Ans}(\tilde{o}^i_k) = L^i\}.
\end{equation}
Finally, we construct the refined SFT dataset $\tilde{\mathcal{D}}$ by aggregating all validated refinements $\tilde{\mathcal{D}}$ across all mathematical queries $\{q^1, \dots, q^n\}$:
\begin{equation}\small
\tilde{\mathcal{D}} = \bigcup_{i=1}^n \tilde{\mathcal{D}}^i.
\end{equation}

\begin{table*}[t]
\centering
\small
\resizebox{\textwidth}{!}{
\begin{tabular}{l|rr|rr|rr|rr|rr}
\hline
{\textbf{Model}} & \multicolumn{2}{c|}{\textbf{AIME24}} & \multicolumn{2}{c|}
{\textbf{AMC23}}& \multicolumn{2}{c|}
{\textbf{Math500}} & \multicolumn{2}{c|}{\textbf{GSM-H}} & \multicolumn{2}{c}{\textbf{Avg.}} \\ 
&Acc@1 & Acc@10 & Acc@1 & Acc@10 & Acc@1 & Acc@10& Acc@1 & Acc@10& Acc@1 & Acc@10 \\ \hline
\multicolumn{11}{c}{\cellcolor{gray!8}{%
  \parbox[c][3.5ex][c]{\linewidth}{\centering \textbf{Qwen3-8B-Instruct}}
}} \\ \hline

Vanilla LLM &20.00	&21.10	&55.00	&53.10	&81.40	&80.75	&57.40	&55.82	&53.45	&52.69 \\ 
Wrong-of-Thought &16.67 & 16.90 & 55.50 & 54.85 & 79.83 & 80.14 & 58.16 & 57.95 & 52.54 & 52.46 \\
SFT (Label) &16.67 & 18.83 & 52.50 & 51.55 & 80.95 & 79.82 & 57.13 & 56.55 & 51.81 & 51.69 \\
SFT (Long-CoT) &23.33	&22.50	&57.50	&56.90	&82.90	&81.40	&59.27	&58.82	&55.75	&54.91 \\ 
\method{} & \textbf{26.67} & \textbf{25.60} & \textbf{62.50} & \textbf{61.85} & \textbf{83.50} & \textbf{82.95} & \textbf{59.83} & \textbf{59.75} & \textbf{58.13} & \textbf{57.54} \\ 
\hline
\multicolumn{11}{c}{\cellcolor{gray!8}{%
  \parbox[c][3.5ex][c]{\linewidth}{\centering \textbf{Qwen2.5-7B-Instruct}}
}} \\ \hline

Vanilla LLM &10.00	&6.20	&47.50	&44.00	&69.80	&69.80	&55.72	&54.34	&45.76	&43.59 \\
Wrong-of-Thought &3.33 & 2.79 & 22.75 & 24.45 & 70.50 & 69.87 & \textbf{59.39} & \textbf{58.91} & 38.99 & 39.01 \\
SFT (Label) &6.67 & 6.55 & 45.00 & 46.25 & 68.85 & 69.35 & 54.31 & 54.09 & 43.71 & 44.06 \\
SFT (Long-CoT) &13.33	&11.35	&50.00	&48.55	&72.10	&70.21	&57.21	&55.85	&48.16	&46.49 \\
\method{} &\textbf{16.67} & \textbf{14.83} & \textbf{55.00} & \textbf{53.51} & \textbf{73.80} & \textbf{72.30} & 58.86 & 58.24 & \textbf{51.08} & \textbf{49.72}\\\hline

\multicolumn{11}{c}{\cellcolor{gray!8}{%
  \parbox[c][3.5ex][c]{\linewidth}{\centering \textbf{Llama3.1-8B-Instruct}}
}}  \\ \hline

Vanilla LLM  &3.33	&2.84	&22.50	&22.50	&43.60	&41.80	&31.23	&32.16	&25.17	&24.83	 \\ 
Wrong-of-thought & 0.00 & 0.00 & 12.50 & 9.95 & 44.00 & 43.98 & 31.92 & 32.97 & 22.11 & 21.73

 \\
SFT (Label) &3.33 & 2.28 & 20.00 & 21.83 & 42.80 & 42.14 & 30.95 & 31.18 & 24.27 & 24.36
 \\
SFT (Long-CoT) &6.67	&4.00	&25.00	&24.50	&44.50	&43.76	&31.83	&32.10	&27.00	&26.09 \\
\method{} &\textbf{10.00}	&\textbf{6.80}	&\textbf{30.00}	&\textbf{27.00}	&\textbf{45.70}	&\textbf{44.78}	&\textbf{33.34}	&\textbf{33.95}	&\textbf{29.76}	&\textbf{28.13} \\ \hline
\end{tabular}}
\caption{Overall Performance. The \textbf{best} result among all methods is highlighted for clarity and emphasis.}
\label{tab:overall-longcot}
\end{table*}

\section{Experimental Methodology}
This section first describes the datasets, evaluation metric, and baselines, followed by the implementation details of our experiments.

\textbf{Dataset.} In our experiments, we randomly sample 10,000 examples from the 
OpenR1-Math-220k dataset~\cite{openr1}, which contains high-quality solutions annotated by DeepSeek-R1~\cite{deepseekai2025deepseekr1incentivizingreasoningcapability}, to construct our training dataset. For evaluation, we select four mathematical problem datasets spanning a range of difficulty levels.
For evaluation, we select five mathematical reasoning benchmarks of varying difficulty, including GSM-Hard~\citep{gao2023pal}, MATH500~\citep{godahewa2021monashtimeseriesforecasting}, AIME24, and AMC23~\citep{omini}. GSM-Hard is an enhanced version that increases the computational difficulty of the GSM8K dataset~\cite{cobbe2021trainingverifierssolvemath}. MATH500 is a benchmark of competition math problems of varying difficulty. AIME and AMC test mathematical problem-solving with arithmetic, algebra, counting, geometry, number theory, probability, and other secondary school math topics.

\textbf{Baselines.} We compare our \method{} model with both zero-shot models and Supervised Fine-Tuning (SFT) models. We first compare with two zero-shot baselines: vanilla LLM and Wrong-of-Thought~\citep{zhang2024wrong}. For the vanilla LLM, we provide the mathematical question $q$ to the model and ask it to directly generate a solution. Additionally, we compare our approach with Wrong-of-Thought~\citep{zhang2024wrong} in our experiments. The model first generates an initial answer, then identifies solution errors using a multi-perspective verifier, and regards these errors as signals to help LLMs prevent the repetition of the same mistakes. Furthermore, we utilize two SFT-based LLMs as baselines, including SFT (Label) and SFT (Long-CoT)~\citep{deepseekai2025deepseekr1incentivizingreasoningcapability}. Following previous work~\citep{wang2024learning}, SFT (Label) and SFT (Long-CoT) fine-tune the LLMs using human-annotated labels and reasoning traces from RLMs.

\textbf{Evaluation Metric.} Following prior studies~\citep{reich2023overcome,zhang2024wrong}, we adopt the Accuracy score as the evaluation metric to determine whether the mathematical QA models correctly generate the ground-truth answers. Specifically, we report Acc@1 and Acc@10, where Acc@1 and Acc@10 represent the average accuracy over one and ten sampled answers at inference time, respectively~\citep{liu2025llmscapablestablereasoning}.

\textbf{Implementation Details.} For all experiments, we employ Qwen2.5-7b-Instruct~\citep{qwen2025qwen25technicalreport}, Qwen3-8b-Instruct~\citep{yang2025qwen3technicalreport} and Llama3.1-8b-Instruct~\citep{grattafiori2024llama3herdmodels} as the backbone models for implementing the mathematical tasks. During training, the learning rate is set to $5 \times 10^{-5}$, with a gradient accumulation step of 8, and each model is trained for 3 epochs. We also use LoRA~\citep{hulora} for efficient training. Additionally, our \method{} model is built on TRL\footnote{\url{https://github.com/huggingface/trl}} and LLaMA Efficient Tuning\footnote{\url{https://github.com/hiyouga/LLaMA-Factory}}. We provide additional experimental details in Appendix~\ref{app:data statistics}, and more details of prompt templates are provided in Appendix~\ref{app:prompt-ECHO}.








\section{Evaluation Result}
In this section, we first evaluate the overall performance of \method{} through the main experience. Next, we conduct ablation studies to investigate the specific contribution of solution error and self-reflection components. Furthermore, we then analyze the effectiveness of distilled models via \method{}. Finally, some case studies are shown in Appendix~\ref{app:case}.

\subsection{Overall Performance}
As shown in Table~\ref{tab:overall-longcot}, we compare the overall performance of \method{} with baseline models across diverse mathematical reasoning tasks. 

Overall, \method{} outperforms all baseline models across datasets, demonstrating its effectiveness. Compared with the prompt-based error-aware learning method Wrong-of-Thought~\cite{zhang2024wrong}, \method{} achieves more than a 5\% improvement, indicating its superior ability to incorporate self-generated errors into model learning. As the evaluation results show, simply injecting solution errors into prompts during inference may degrade LLM performance. This degradation likely occurs because such errors act as noise that interferes with the reasoning process~\cite{wang2024rat}. In contrast to Wrong-of-Thought, \method{} provides a more effective mechanism for guiding LLMs to avoid repeating similar solution errors by fine-graining long-form CoT traces as supervision signals and then fine-tuning student models.

Regarding different SFT strategies, compared with the Vanilla model, SFT (Label) yields nearly identical performance, while SFT (Long-CoT) achieves noticeable improvements. This suggests that the reasoning traces derived from the CoT outputs of RLMs, such as DeepSeek-R1~\cite{deepseekai2025deepseekr1incentivizingreasoningcapability}, provide richer reasoning patterns that can better guide SLMs to imitate during the SFT process. Building upon \method{}, the supervision signals are further refined by the student models themselves through self-reflection and the incorporation of self-identified errors, resulting in an additional 2\% performance gain. These results further demonstrate the effectiveness of \method{} in providing more tailored and effective supervision for student models. Furthermore, to assess generalization, we evaluate \method{} on different backbone models, including Qwen2.5-7B, Qwen3-8B, and LLaMA3.1-8B. The results consistently show notable improvements across all backbones, demonstrating the generalization ability of \method{} in enhancing mathematical reasoning performance.

\begin{table*}[t]
\centering
\small
\resizebox{\textwidth}{!}{
\begin{tabular}{l|rr|rr|rr|rr|rr}
\hline
{\textbf{Model}} & \multicolumn{2}{c|}{\textbf{AIME24}} & \multicolumn{2}{c|}
{\textbf{AMC23}}& \multicolumn{2}{c|}
{\textbf{Math500}} & \multicolumn{2}{c|}{\textbf{GSM-H}} & \multicolumn{2}{c}{\textbf{Avg.}} \\ 
~ &Acc@1 & Acc@10 & Acc@1 & Acc@10 & Acc@1 & Acc@10 & Acc@1 & Acc@10 & Acc@1 & Acc@10  \\ \hline

\multicolumn{11}{c}{\cellcolor{gray!8}{%
  \parbox[c][3.5ex][c]{\linewidth}{\centering \textbf{Qwen3-8B-Instruct}}
}}  \\ \hline
SFT (Long-CoT) &23.33	&22.50	&57.50	&56.90	&82.90	&81.40	&59.27	&58.82	&55.75	&54.91 \\ 
\method{} & \textbf{26.67} & \textbf{25.60} & \textbf{62.50} & \textbf{61.85} & \textbf{83.50} & \textbf{82.95} & \textbf{59.83} & \textbf{59.75} & \textbf{58.13} & \textbf{57.54} \\ 

w/o Solution Error	&26.67	&24.95	&60.00	&58.25	&83.15	&82.10	&59.27	&59.13	&57.27	&56.11 \\
w/o Self-Reflection	&20.00	&20.90	&57.50	&56.50	&82.75	&81.25	&58.86	&58.34	&54.78	&54.25  \\ \hline

\multicolumn{11}{c}{\cellcolor{gray!8}{%
  \parbox[c][3.5ex][c]{\linewidth}{\centering \textbf{Qwen2.5-7B-Instruct}}
}} \\ \hline

SFT (Long-CoT) &13.33	&11.35	&50.00	&48.55	&72.10	&70.21	&57.21	&55.85	&48.16	&46.49 \\

\method{} & \textbf{16.67} & \textbf{14.83} & \textbf{55.00} & \textbf{53.51} & \textbf{73.80} & 72.30 & \textbf{58.86} & \textbf{58.24} & \textbf{51.08} & \textbf{49.72} \\

w/o Solution Error	&16.67	&10.00	&52.50	&52.90	&73.10	&\textbf{72.90}	&58.24	&58.20	&50.13	&48.50 \\
w/o Self-Reflection	&10.00	&5.83	&50.00	&47.86	&70.20	&69.21	&57.35	&56.55	&46.89	&44.86 \\ \hline

\multicolumn{11}{c}{\cellcolor{gray!8}{%
  \parbox[c][3.5ex][c]{\linewidth}{\centering \textbf{Llama3.1-8B-Instruct}}
}} \\ \hline

SFT (Long-CoT) &6.67	&4.00	&25.00	&24.50	&44.50	&43.76	&31.83	&32.10	&27.00	&26.09 \\

\method{} & \textbf{10.00} & \textbf{6.80} & \textbf{30.00} & \textbf{27.00} & \textbf{45.70} & \textbf{44.78} & \textbf{33.34} & \textbf{33.95} & \textbf{29.76} & \textbf{28.13}

 \\ 

w/o Solution Error	&6.67	&5.50	&30.00	&26.50	&45.40	&44.34	&33.13	&32.65	&28.80	&27.25 \\
w/o Self-Reflection	&6.67	&4.80	&27.50	&25.50	&43.90	&43.10	&32.91	&31.95	&27.75	&26.34 \\ \hline
\end{tabular}}
\caption{Ablation Study. We evaluate the performance of distilled models trained with different strategies, excluding the Solution Error and Self-Reflection components, to demonstrate the contribution of each component in \method{}.}
\label{tab:ablation-longcot}
\end{table*}


\subsection{Ablation Study}
As shown in Table~\ref{tab:ablation-longcot}, we conduct an ablation study to isolate the individual effects of the solution error and self-reflection components in \method{}. We compare several ablated variants: \method{} w/o Solution Error removes the injected erroneous solutions during CoT data refinement, while \method{} w/o Self-Reflection directly uses the long-form CoTs as supervisions without refinement and incorporates solution errors into the prompt during training.

Overall, both Self-Reflection and Solution Error are effective in improving the quality of long-form CoT based supervision signals. Specifically, compared with the SFT (Long-CoT) model, \method{} w/o Solution Error achieves over a 1\% improvement, demonstrating the effectiveness of the self-reflection mechanism. This mechanism enables the student model to refine long-form CoTs by itself, thereby narrowing the gap between the supervision signal and the capability of the student model.
Moreover, the \method{} w/o Self-Reflection model also outperforms SFT (Long-CoT), indicating that incorporating self-generated solution errors during the SFT process can guide the student model to avoid repeating potential mistakes.
Further, comparing \method{} w/o Solution Error with the full \method{}, we observe additional gains in reasoning performance, suggesting that considering self-made errors during self-reflection brings complementary benefits. These findings highlight that the advantage of incorporating solution errors can be extended to the CoT refinement process, providing higher-quality and more tailored supervision signals for the student model.

\subsection{The Effectiveness of Distilled Models via Different Training Strategies\label{sec:5.3}}
\begin{figure}[t]
  \centering
    \subfigure[Entropy During Training.\label{fig:entropy}]{\includegraphics[width=0.23\textwidth]{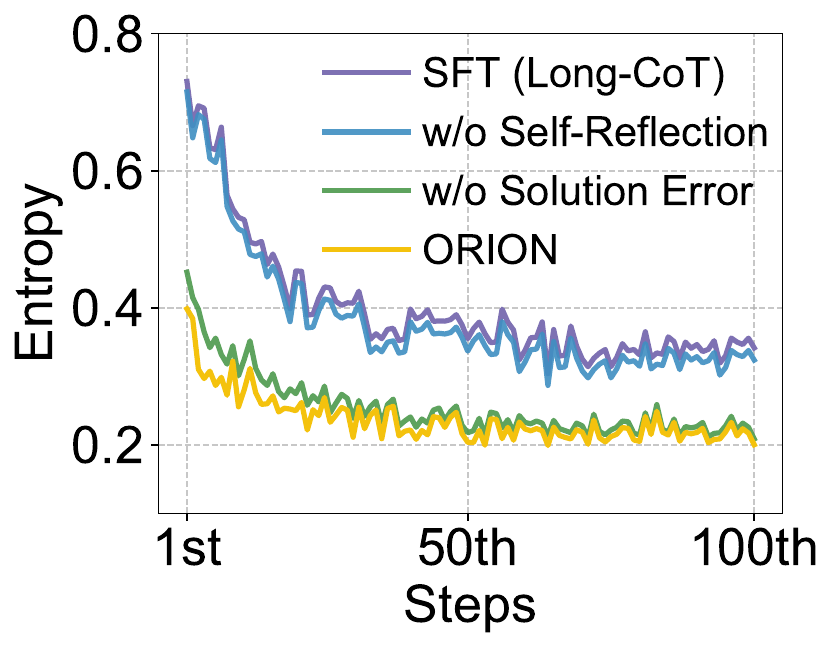}}
    \subfigure[Response Length.]{\includegraphics[width=0.23\textwidth]{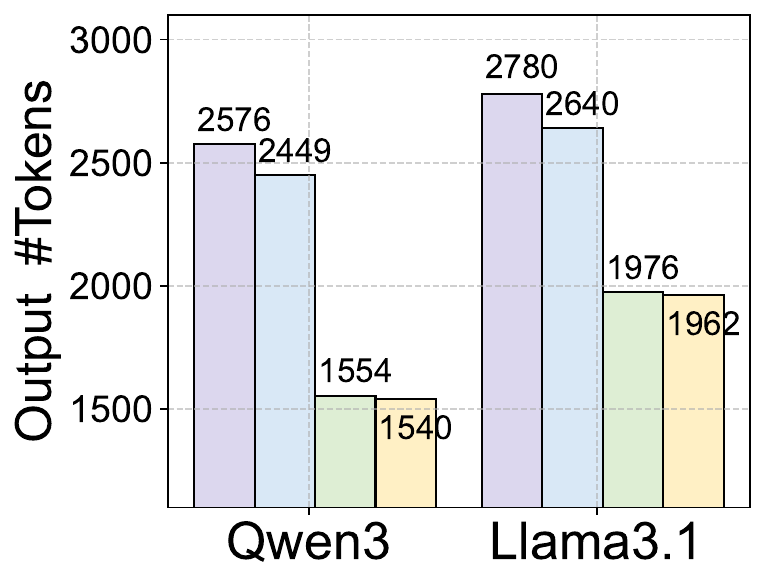}\label{fig:evaluate-error-aware:length-of-response}}
  \subfigure[Perplexity Score.\label{fig:evaluate-error-aware:ppl}]
  {\includegraphics[width=0.23\textwidth]{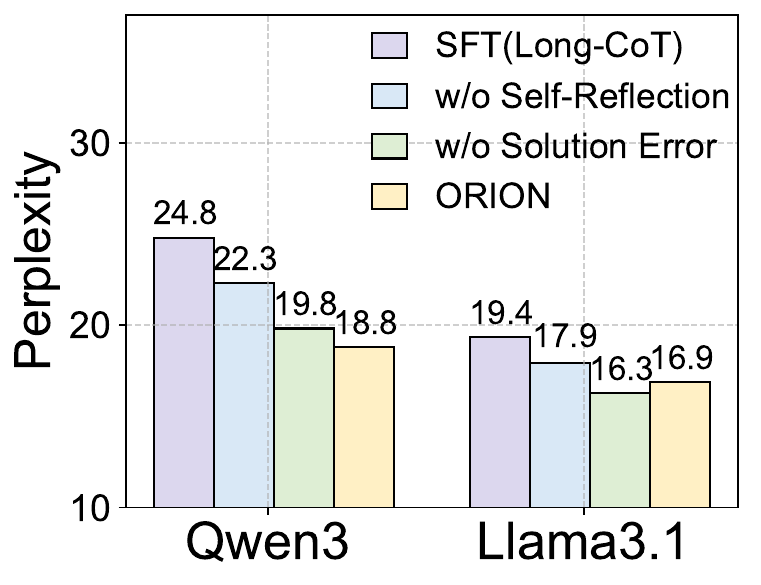}}
    \subfigure[GPT Preference Rate.\label{fig:evaluate-error-aware:gpt_score}]{\includegraphics[width=0.23\textwidth]{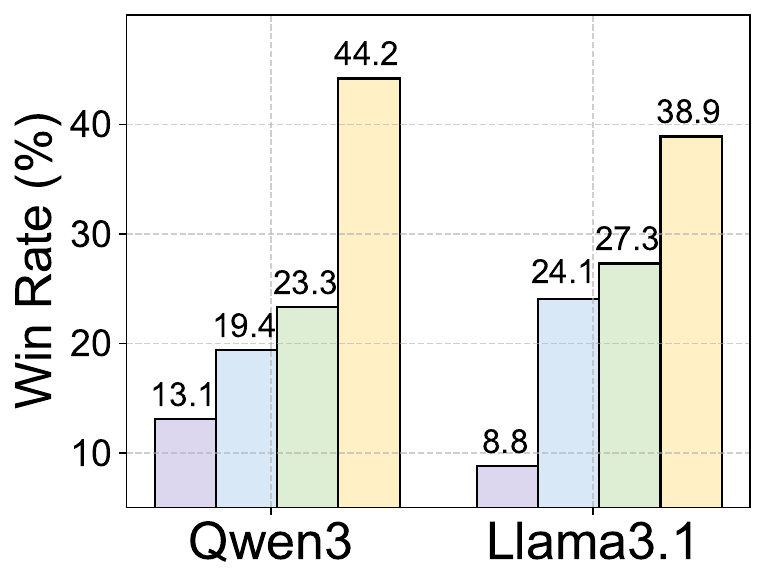}}

\caption{Performance of Distilled Models Optimized with Different Training Strategies. We first report the entropy scores during distillation under various strategies (Figure~\ref{fig:entropy}). We then present the response lengths generated by the distilled models (Figure~\ref{fig:evaluate-error-aware:length-of-response}). Finally, we evaluate the quality of CoT and final responses using both Vanilla LLMs and GPT-4 as judges (Figure~\ref{fig:evaluate-error-aware:ppl} and Figure~\ref{fig:evaluate-error-aware:gpt_score}, respectively).}\label{fig:evaluate-cot}
\end{figure}

In this section, we evaluate the training stability and effectiveness of the distilled LLMs, as shown in Figure~\ref{fig:evaluate-cot}. Three models, SFT (Long-CoT), \method{} w/o Solution Error, and \method{} w/o Self-Reflection, are used as baselines for comparison.

\textbf{Training Stability.}
We further analyze the training stability of different strategies by plotting the entropy scores in Figure~\ref{fig:entropy}. Compared with both SFT (Long-CoT) and \method{} w/o Self-Reflection, the \method{} and \method{} w/o Solution Error models exhibit significantly lower entropy scores during training, demonstrating their effectiveness in promoting a more stable and learnable training process. This phenomenon further confirms that the self-reflection mechanism benefits the student model’s learning process during SFT.
Moreover, when incorporating self-generated solution errors, the entropy scores decrease slightly, suggesting that these errors can be effectively utilized to refine long-form CoTs and better align them with the learning capacity of student models. Overall, these results highlight the crucial role of narrowing the gap between teacher-provided supervision and the learning ability of student models in SFT-based distillation.

\textbf{Effectiveness of Distilled Models.}
To evaluate the distilled models trained with different strategies, we analyze both the length (Figure~\ref{fig:evaluate-error-aware:length-of-response}) and quality (Figures~\ref{fig:evaluate-error-aware:ppl} and \ref{fig:evaluate-error-aware:gpt_score}) of the generated CoTs.

As shown in Figure~\ref{fig:evaluate-error-aware:length-of-response}, we prompt each distilled model to generate reasoning trajectories and measure their average lengths. The results indicate that both \method{} and \method{} w/o Solution Error produce significantly shorter reasoning trajectories compared with other models. This finding suggests that the self-reflection mechanism helps eliminate rigid or redundant reasoning patterns, enabling the student model to learn more concise and efficient reasoning processes. Consequently, it helps mitigate the overthinking phenomenon~\cite{chen2024not} and reduces inference costs.
Next, we assess the quality of the generated CoTs. As shown in Figure~\ref{fig:evaluate-error-aware:ppl}, we collect the generated reasoning traces and use the corresponding LLMs as judges to compute the perplexity scores with respect to the ground-truth answers. The results show that \method{} achieves the lowest perplexity among all models, demonstrating that its generated CoTs better align with the true reasoning process and can effectively guide zero-shot LLMs toward correct answers.
Furthermore, as shown in Figure~\ref{fig:evaluate-error-aware:gpt_score}, we employ GPT-4 as an additional evaluator to assess the generated responses based on criteria such as correctness and clarity. Specifically, GPT-4 is prompted to select the best responses among all four models using the prompt templates provided in Appendix~\ref{app:prompt-score}. The evaluation results demonstrate that \method{} achieves twice the win rate of other models, highlighting its effectiveness in helping to distill more capable and reliable reasoning models.

\begin{figure}[t]
  \centering
  \subfigure[Error Type Distribution.\label{fig:category:type}]{\includegraphics[width=0.22\textwidth]{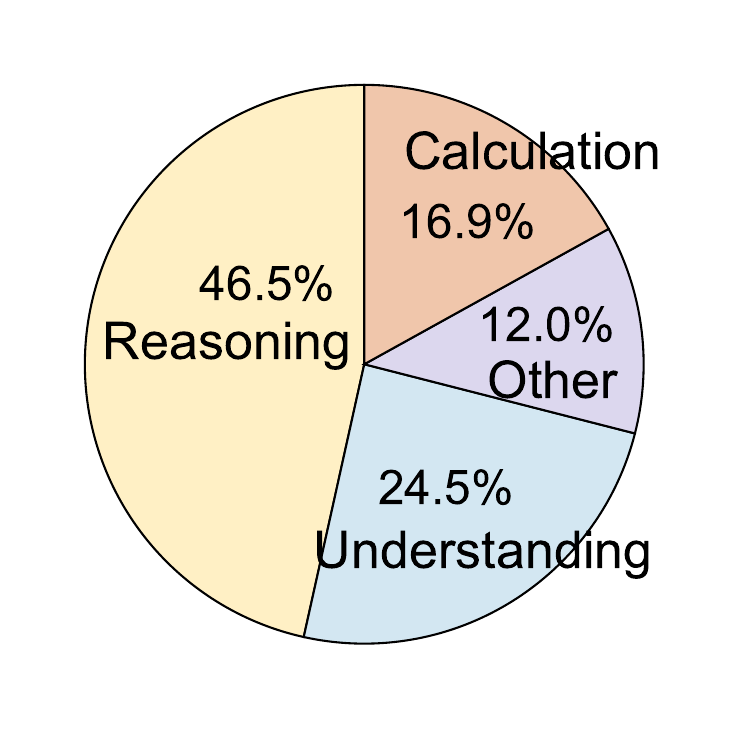}}
  \subfigure[Reasoning Error.\label{fig:category:reasoning}]{\includegraphics[width=0.23\textwidth]{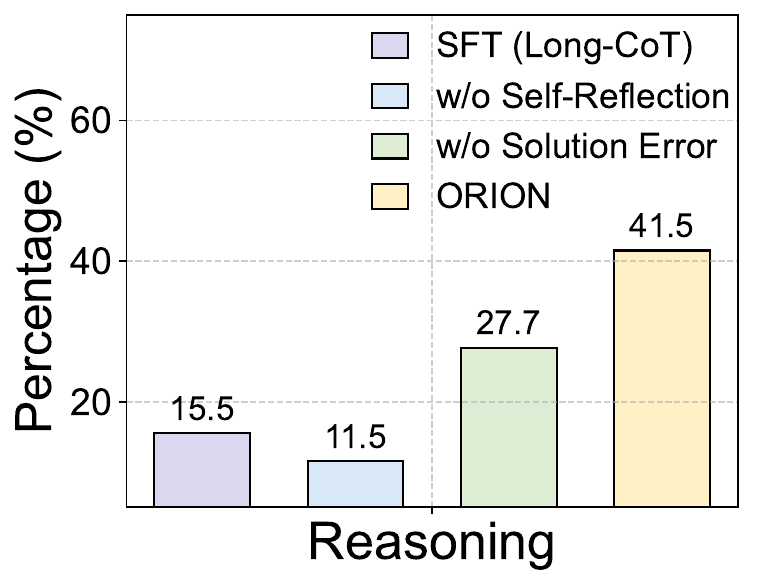}}
  \subfigure[Calculating Error.\label{fig:category:calculation}]{\includegraphics[width=0.22\textwidth]{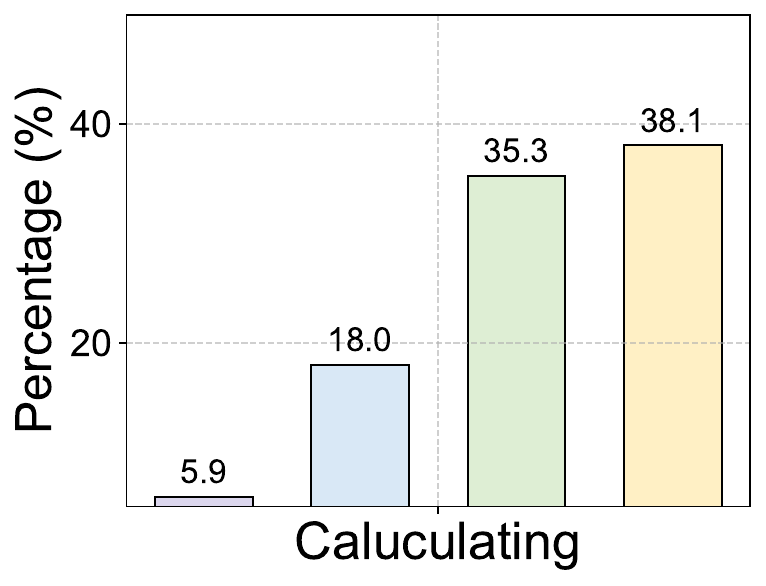}}
  \subfigure[Understanding Error.\label{fig:category:understanding}]{\includegraphics[width=0.22\textwidth]{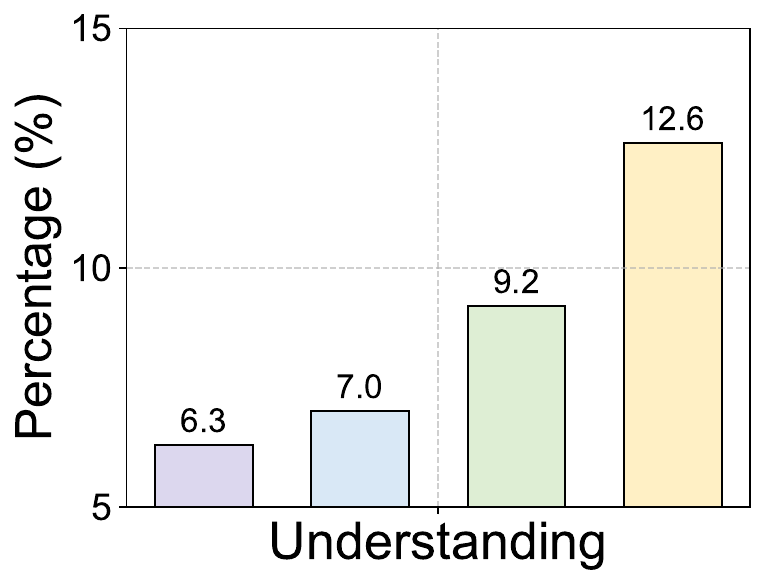}}

  \caption{Analysis of Distilled Models on Different Error Types. All experiments are based on the Qwen3-8B model. Figure~\ref{fig:category:type} shows the distribution of distinct error types encountered by vanilla LLMs, while Figures~\ref{fig:category:reasoning}, \ref{fig:category:calculation}, and \ref{fig:category:understanding} present the corresponding correction rates for each error type.}
  \label{fig:category}
\end{figure}

\subsection{The Performance of Distilled Models on Different Error Types}\label{app:category}
To further validate the effectiveness of our targeted error exposure strategy, we conduct a detailed analysis of error types and their corresponding correction outcomes, as illustrated in Figure~\ref{fig:category}. Specifically, we first prompt the vanilla LLM to generate responses on the entire test set and sample 500 instances that are incorrectly answered for further analysis. These sampled instances are then categorized into four distinct error types, reasoning, calculation, and understanding errors, using the prompt template provided in Appendix~\ref{app:prompt-type}.

As shown in Figure~\ref{fig:category:type}, we present the distribution of error categories produced by the vanilla LLM. The analysis reveals that reasoning errors constitute the majority of all mistakes, indicating that vanilla LLMs struggle most with reasoning-intensive problems. This observation underscores the importance of enhancing the reasoning ability of student models in mathematical problem-solving, such as distilling reasoning capabilities from LRMs.
We then evaluate the effectiveness of different distilled models in correcting each error type, as shown in Figures~\ref{fig:category:reasoning}, \ref{fig:category:calculation}, and \ref{fig:category:understanding}. For reasoning errors, \method{} achieves the highest correction rate of approximately 41\%, demonstrating its strong ability to help the student model overcome the most common error type. This improvement can be attributed to the refined supervision signals provided by the self-reflection mechanism, which guide the model to internalize effective reasoning patterns. Furthermore, compared with \method{} w/o Solution Error, \method{} achieves significantly higher correction rates on both reasoning and understanding errors, indicating that the inclusion of self-generated solution errors further enhances the model’s ability to refine supervision signals and avoid repeating similar reasoning or comprehension mistakes.

\section{Conclusion}
This paper proposes the Error-Aware Self-Reflection (err\textbf{O}r-aware self-\textbf{R}eflect\textbf{ION}, \method{}) method, which distills the strong reasoning capabilities of teacher models into a student model implemented as SLMs via SFT. \method{} leverages the student model's own self-reflection to improve the long-form CoTs generated by Long Reasoning Models (LRMs), thereby generating more effective supervision signals. Experimental results show that \method{} outperforms all baselines across multiple mathematical tasks.


\section*{Limitation}
Although \method{} has proven effective in constructing higher-quality SFT data, it still relies on powerful closed-source reasoning language models to generate long-form reasoning chains, which incur significant computational costs. In addition, \method{} only utilizes long-form CoTs from Deepseek-R1 as the training target and has not explored the effectiveness of using other closed-source long-form reasoning models as teacher models, primarily due to the high time and computational costs involved in generating reasoning data with closed-source models.


%

\bibliography{arr2026}
\clearpage

\appendix
\section{Appendix}
\subsection{License}
We provide the licenses for the datasets we use. OpenR1-Math-220k and MATH500 are licensed under the Apache License 2.0. AIME24 and AMC23 are not currently labeled with license types. GSM-Hard uses the MIT license. All of these licenses and agreements allow their data for academic use.

\subsection{Additional Implementation Details}\label{app:data statistics}
In this section, we provide the detailed data statistics in ORION. We build our training set by using 10,000 examples randomly sampled from OpenR1-Math-220k~\cite{openr1} as seed data and then expand it to form our final training data. For evaluation, we use the GSM-Hard dataset containing 1,319 examples, MATH500 with 500 examples, AIME24 with 30 examples, and AMC23 with 40 examples. All statistics are shown in Table~\ref{tab:dataset}.

\begin{table}[t]
\centering
\small
\begin{tabular}{l|r|r}
\hline
\textbf{Dataset} & \textbf{Train}  & \textbf{Test} \\  
\hline
OpenR1-Math-220k~\shortcite{openr1} & 10,000  & - \\
GSM-Hard~\shortcite{gao2023pal} & -  & 1,319 \\                    
MATH500~\shortcite{godahewa2021monashtimeseriesforecasting} & - & 500 \\  
AIME24~\shortcite{omini} & - & 30 \\  
AMC23~\shortcite{omini} & - & 40\\  
\hline
\end{tabular}
\caption{Data Statistics.}
\label{tab:dataset}
\label{table1:datasets}  
\end{table}




\subsection{Prompt Templates Used in \method{}}\label{app:prompt-ECHO}
\begin{figure*}[t]
    \centering
    \includegraphics[width=0.9\linewidth]{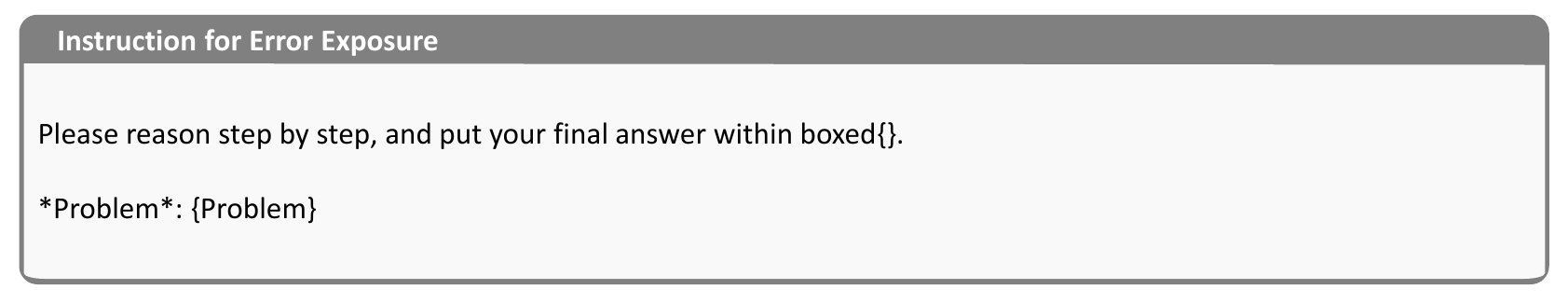}
    \caption{The Prompt Templates Used for Error Exposure.}
    \label{fig:prompt-error-exposure}
\end{figure*}

\begin{figure*}[t]
    \centering
    \includegraphics[width=0.9\linewidth]{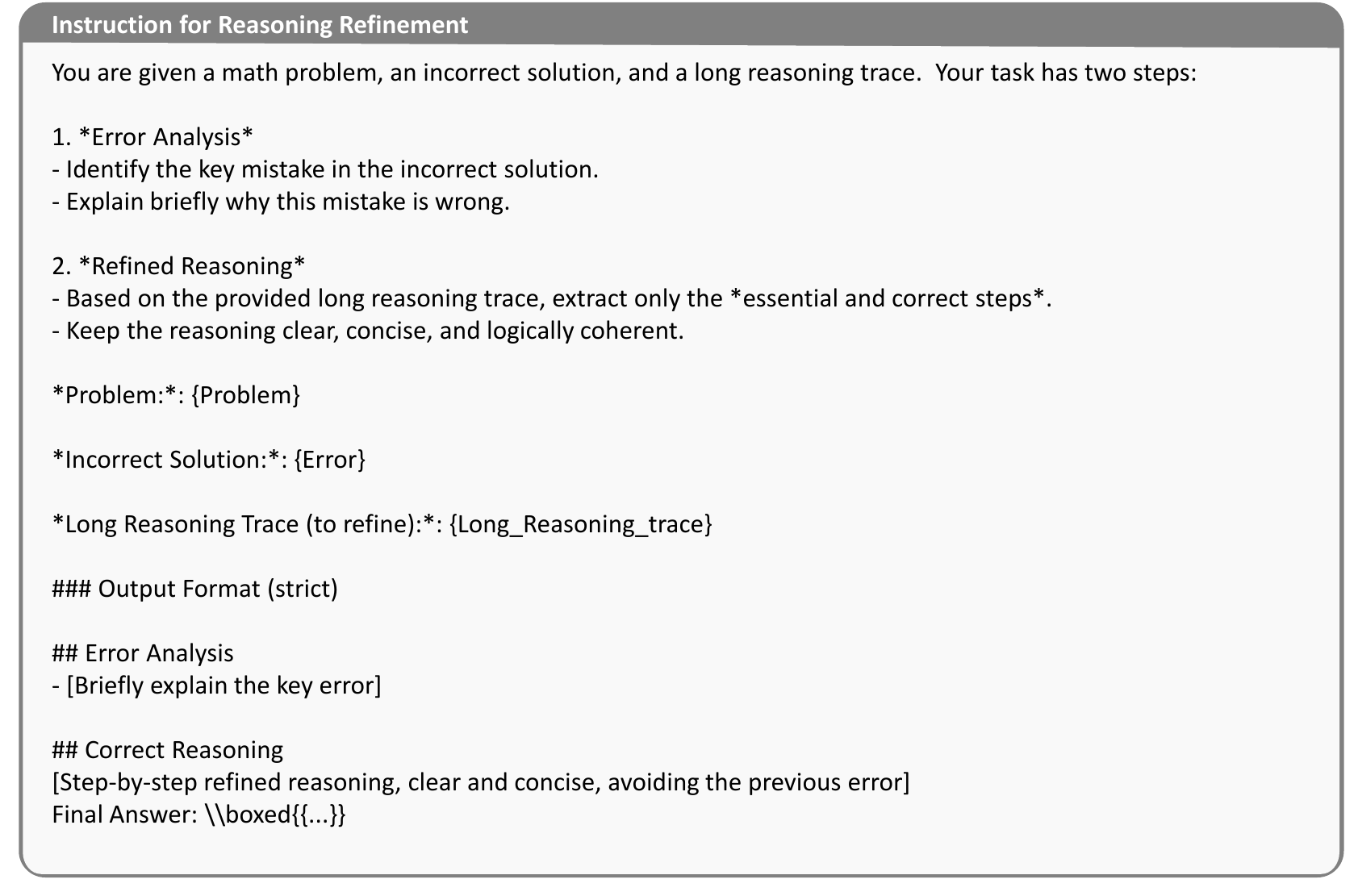}
    \caption{The Prompt Templates Used for Reasoning Refinement.}
    \label{fig:prompt-reasoning-refinement}
\end{figure*}

\begin{figure*}[t]
    \centering
    \includegraphics[width=0.9\linewidth]{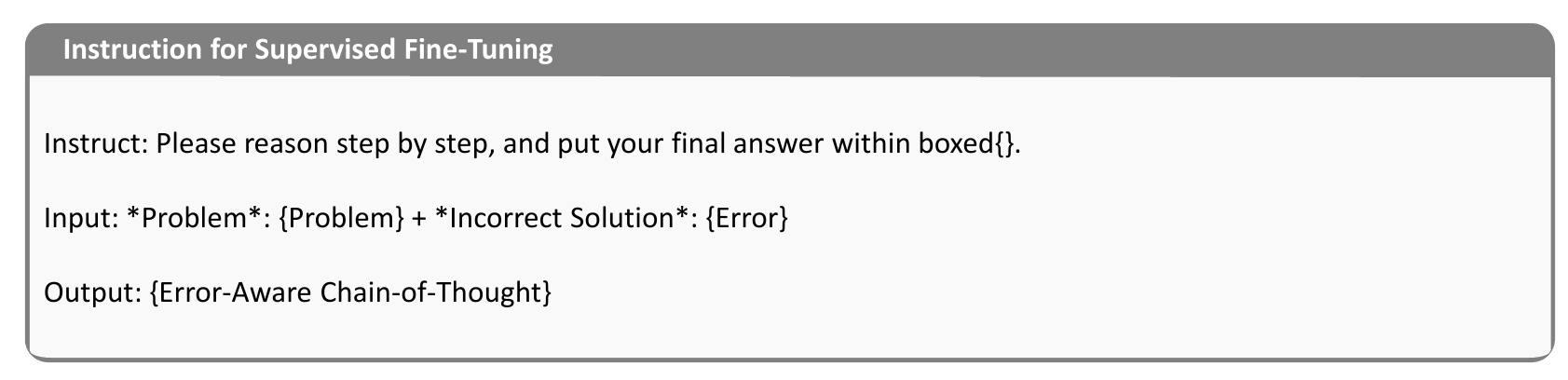}
    \caption{The Prompt Templates Used for Supervised Fine-Tuning.}
    \label{fig:prompt-sft}
\end{figure*}

We detail the prompts employed across different stages of \method{}. As shown in Figure~\ref{fig:prompt-error-exposure}, the Error Exposure stage uses prompts that instruct the model to solve problems, from which incorrect cases are collected. These errors are then utilized in the Self-Reflection stage, where the prompt templates in Figure~\ref{fig:prompt-reasoning-refinement} instruct the model to identify and correct solution errors while refining the long-form CoTs. Finally, during the Supervised Fine-Tuning stage, the prompt in Figure~\ref{fig:prompt-sft} leverages the refined trajectories with their corresponding solution errors as training supervision.

\subsection{Prompt Templates Used for Evaluating the Quality of \method{}}\label{app:prompt-score}
\begin{figure*}[t]
    \centering
    \includegraphics[width=0.91\textwidth]{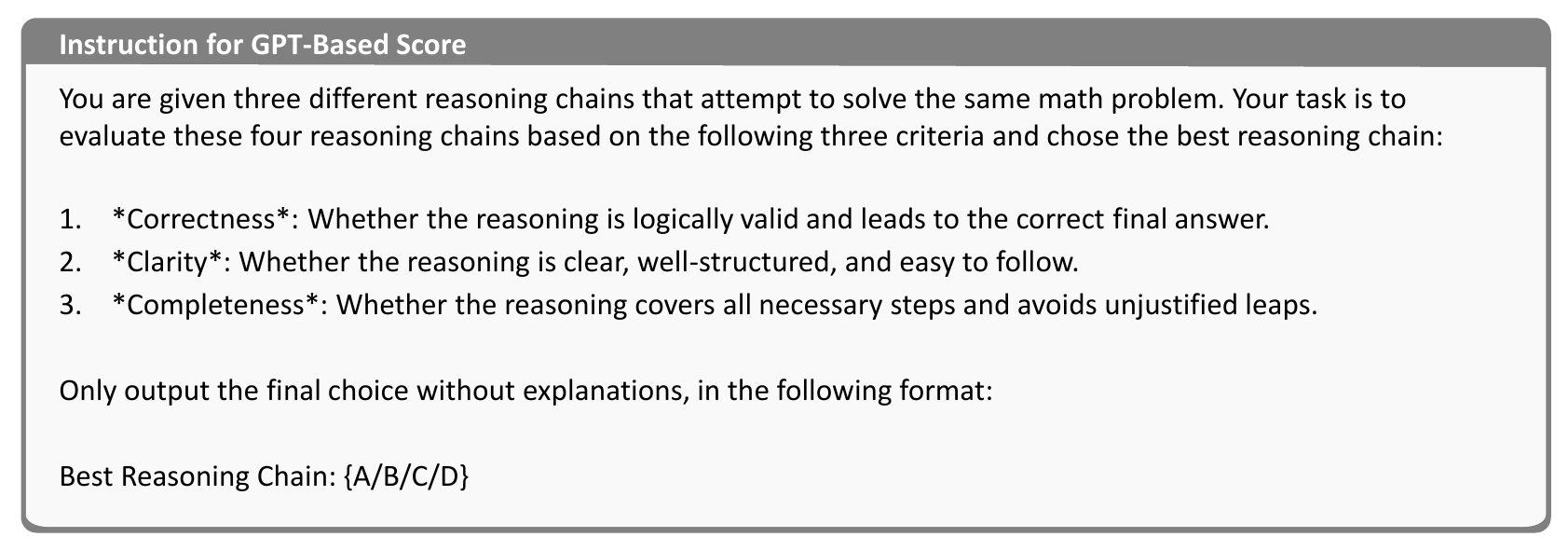}
    \caption{The Prompt Templates Used in GPT Preference Rate.}
    \label{fig:prompt-gptscore}
\end{figure*}

As shown in Figure~\ref{fig:prompt-gptscore}, we present the prompt templates used to evaluate the quality and educational value of the CoTs synthesized by four methods: SFT (Long-CoT), \method{} w/o Solution Error, \method{} w/o Self-Reflection, and \method{}. The evaluation is conducted using GPT-4~\cite{openai2024gpt4technicalreport} as the evaluator, which assesses each generated reasoning trace based on its clarity, correctness, and pedagogical usefulness. We employ consistent prompt templates across all methods to ensure fair comparison and reproducibility.

\subsection{Prompt Templates Used for Solution Error Categorization}\label{app:prompt-type}
\begin{figure*}[t]
    \centering
    \includegraphics[width=0.9\linewidth]{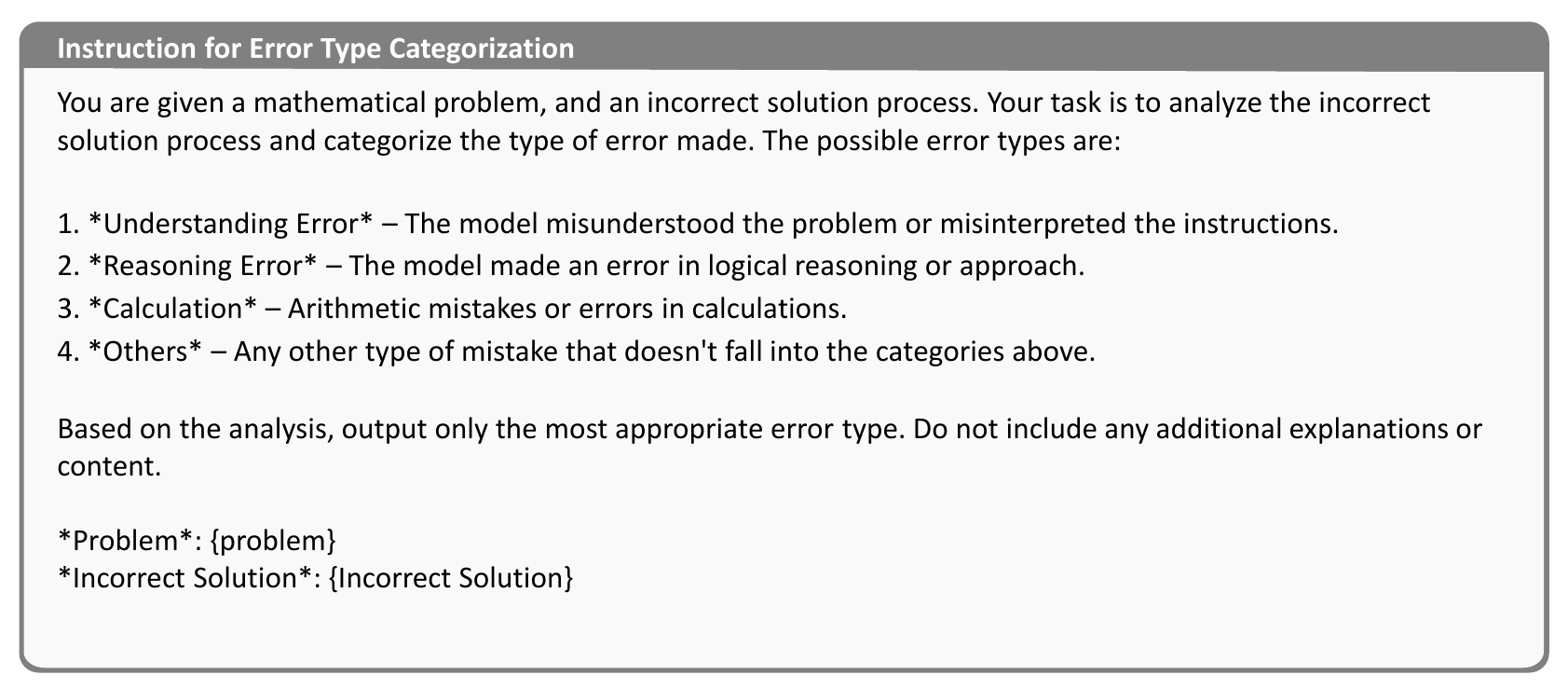}
    \caption{The Prompt Templates Used for Error Type Categorization.}
    \label{fig:prompt-category}
\end{figure*}

As shown in Figure~\ref{fig:prompt-category}, we present the prompt templates used to classify the solution errors generated by GPT-4~\cite{openai2024gpt4technicalreport}. The prompt instructs the evaluator model to classify each incorrect response into one of four predefined categories: Reasoning Error, Understanding Error, Calculation Error, or Other.

\subsection{Case Study}\label{app:case}

In this section, we present a detailed case study across two tables to intuitively demonstrate the core process and the resulting effectiveness of \method{}. As shown in Table~\ref{case-study:casestudy-refinement}, we illustrate the internal mechanism of our method. It presents the error exposure and self-reflection process. Building on this, Table~\ref{case-study:casestudy-different-method} contrasts the final response generated by the \method{} with those from baseline methods, showing the superior quality and clarity of its output.

First, we examine a relatively simple math problem to analyze the behavior of \method{} during the error exposure phase. Despite the low comlexity of the problem, sampling multiple responses reveals a diverse range of errors, including a hallucination for general concepts (``lower class soldiers (8) is more than the available (10)''), misunderstandings of the problem statement (``exceeding their respective limits'') and basic computational mistakes (``C(8,10)=90''). Therefore, we argue that it is crucial to expose as many diverse erroneous responses as possible for each problem and to use them as guidance to avoid similar pitfalls in the final reasoning. Then, we present a complete example of the Self-Reflection process. Given a mathematical problem, an incorrect solution, and a standard long-form CoT from an RLM, the model generates an Error-Aware Chain-of-Thought that not only analyzes the key mistakes in the solution error but also refines the standard reasoning trace to make it better aligned with the model's own capacity.

As shown in Table~\ref{case-study:casestudy-different-method}, we contrast the final responses to illustrate the superior reasoning capability of the full \method{} model. The \method{} w/o Self-Reflection provides a direct solution but lacks any analysis of potential pitfalls, resulting in an incorrect response. The \method{} w/o Solution Error produces an analysis of possible error; however, due to the absence of error examples in its training, it ultimately fails to deliver a correct answer. In stark contrast, the complete \method{} model exhibits self-awareness by first identifying a potential failure point, noting that the solution could be flawed (``because the expression for n was not properly simplified or checked for integer solutions.''). Building on this crucial analysis, it then presents a clear, refined solution, demonstrating its ability to anticipate and preempt common errors.


\begin{table*}[t]
\centering
\small
\renewcommand\arraystretch{1.2}
\setlength{\fboxsep}{1.0pt}
\resizebox{\linewidth}{!}{ 
\begin{tabular}{p{0.88\textwidth}}
\hline
\rowcolor{gray!8}{\textbf{Case 1:} The Case of
Error Exposure via Response Sampling.}\\

\hline
\textbf{Question:} The Greek army contained two types of soldiers: the upper class and the lower class soldiers. If there were a total of 5 upper class soldiers, and 10 lower class soldiers in a certain part of Athens, and the battle of Thermopylae demands a force of 4 upper class soldiers and 8 lower class soldiers, how many different battalions can be sent?
\\
\textbf{Ground-Truth:} 225
\\
\textbf{Error Response[0]:} To solve this problem, we need to determine how many different battalions can be formed given the constraints on..., Since the requirement for \textbf{\textcolor[rgb]{0.7,0.3,0.3}{lower class soldiers (8) is more than the available (10)}}, it is impossible to...\textbf{\textcolor[rgb]{0.7,0.3,0.3}{So the final answer is boxed\{0\}.}}
\\
\textbf{Error Response[1]:} To solve this problem, we need to determine how many different battalions can be formed given the constraints on..., Since both the upper class and lower class soldiers allow us to form exactly one battalion without ...\textbf{\textcolor[rgb]{0.7,0.3,0.3}{exceeding their respective limits...So the final answer is boxed\{1\}.}}
\\
\textbf{Error Response[2]:} To solve this problem, we need to determine how many different battalions can be formed given the constraints on the number of upper class and lower class soldiers available... For lower class soldiers, the number of ways to choose 8 from 10 is mistakenly calculated as \textbf{\textcolor[rgb]{0.7,0.3,0.3}{C(10,8)=90}}, Then, the total number of battalions is 
$5 \times 90 = 450$. \textbf{\textcolor[rgb]{0.7,0.3,0.3}{So the final answer is boxed\{450\}.}}
\\
\hline
\rowcolor{gray!8}{\textbf{Case 2:} The Case of Reasoning Refinement via Error-Aware Self-Reflection.}
\\
\hline
\textbf{Question:} Let \$a, b, c\$ be distinct numbers such that the equations $x^2 + ax + 1 = 0$ and $x^2 + bx + c = 0$ have a common real root, and the equations $x^2 + x + a = 0$ and $x^2 + cx + b = 0$ also have a common real root. Compute the sum $a + b + c$. 
\\
\textbf{Ground-Truth:} -3
\\
\textbf{Error Response:} To solve for \(a + b + c\), we need to analyze the given conditions step by step. Let the common real root of the equations \(x^2 + ax + 1 = 0\) and \(x^2 + bx + c = 0\) be \(r\); substituting \(r\) into both equations gives \(r^2 + ar + 1 = 0\) (1) and \(r^2 + br + c = 0\) (2). Subtracting (1) from (2) yields \((b - a)r + (c - 1) = 0\), so \(r = \frac{1 - c}{b - a}\) (3). Similarly, let the common real root of \(x^2 + x + a = 0\) and \(x^2 + cx + b = 0\) be \(s\); substituting \(s\) gives \(s^2 + s + a = 0\) (4) and \(s^2 + cs + b = 0\) (5). Subtracting (4) from (5) yields \((c - 1)s + (b - a) = 0\), so \(s = \frac{a - b}{c - 1}\) (6). Since \(r = s\), equating (3) and (6) gives \(\frac{1 - c}{b - a} = \frac{a - b}{c - 1}\); cross-multiplying gives \((1 - c)(c - 1) = (b - a)^2\), or \(-(c - 1)^2 = (b - a)^2\). Since \(a, b, c\) are distinct, \((b - a)^2 \neq 0\), so \((c - 1)^2 = (b - a)^2\), meaning \(c - 1 = \pm (b - a)\). Considering \(c = b - a + 1\) and substituting back, we find \(r = -1\) from (3). Substituting \(r = -1\) into \(x^2 + x + a = 0\) gives \(1 - 1 + a = 0\), so \(a = 0\). Then \(c = b + 1\); checking consistency with \(x^2 + cx + b = 0\) confirms validity, giving \(b = -1\) and \(c = 0\). \textbf{\textcolor[rgb]{0.7,0.3,0.3}{Thus, the final answer is boxed\{-1\}.}}
\\
\textbf{Reasoning Refinement:} 
*Error Analysis* : 
The incorrect solution \textbf{\textcolor[rgb]{0.13,0.67,0.8}{incorrectly assumed that the common roots \( \alpha \) and \( \beta \) must be the same.}} leading to a contradiction with the requirement that \( a, b, c \) are distinct. Specifically, it derived \( \beta = \frac{1 - c}{a - b} \) and \( \alpha = \frac{c - 1}{b - a} \), then concluded \( \alpha = \beta \), leading to \( \beta = 1 \), \( \alpha = 1 \), \( a = -2 \), and \( b = -1 - c \). This made \( a, b \) dependent on \( c \), conflicting with the distinctness condition.
\\
*Correct Reasoning* :
Let’s start by carefully defining the problem to avoid confusion. We need to find \( a + b + c \) where \( a, b, c \) are distinct numbers, with two pairs of quadratic equations sharing common real roots. First, let’s clarify which roots are common—they don’t have to be the same across both pairs, \textbf{\textcolor[rgb]{0.13,0.67,0.8}{which was a key mistake in the incorrect solution.}}  

First, define the common roots properly: Let the common real root of \( x^2 + ax + 1 = 0 \) and \( x^2 + bx + c = 0 \) be \( \alpha \); this means \( \alpha \) satisfies both equations. Separately, let the common real root of \( x^2 + x + a = 0 \) and \( x^2 + cx + b = 0 \) be \( \beta \), so \( \beta \) satisfies these two equations. Note that \( \alpha \) and \( \beta \) might not be equal—this distinction is crucial.  

Next, use Vieta’s Formulas to relate the roots and coefficients of each quadratic equation. For a quadratic equation \( x^2 + px + q = 0 \), the sum of roots is \( -p \) and the product is \( q \)...


Now, let’s connect these equations using the fact that \( \alpha \) and \( \beta \) satisfy their respective pairs of equations. For \( \alpha \) in both quadratics, subtracting the equations gives \( (b - a)\alpha + (c - 1) = 0 \implies \alpha = \frac{c - 1}{b - a} \). For \( \beta \), subtracting its pair gives \( (1 - c)\beta + (a - b) = 0 \implies \beta = \frac{b - a}{c - 1} \). Notice that \( \alpha \cdot \beta = 1 \)—interesting, but let’s focus on \( \alpha \).  

Let’s simplify by testing specific values for \( \alpha \). Suppose \( \alpha = 1 \): substituting into \( x^2 + ax + 1 = 0 \) gives \( 1 + a + 1 = 0 \implies a = -2 \). For \( x^2 + bx + c = 0 \), \( 1 + b + c = 0 \implies b = -1 - c \).  

Now check the second pair with \( a = -2 \): \( x^2 + x - 2 = 0 \) factors to \( (x + 2)(x - 1) = 0 \), so \( \beta \) is \( 1 \) or \( -2 \). If \( \beta = 1 \), substituting into \( x^2 + cx + b = 0 \) gives \( 1 + c + b = 0 \), which holds since \( b = -1 - c \). If \( \beta = -2 \), we get \( c = 1 \) and \( b = -2 \), but this makes \( a = b \), violating distinctness—so \( \beta = 1 \) is valid.  

Ensuring \( a, b, c \) are distinct (e.g., \( c \neq 1, -2 \)), the sum is \( a + b + c = -2 + (-1 - c) + c = -3 \).

\textbf{\textcolor[rgb]{0.13,0.67,0.8}{Thus, the final answer is boxed\{-3\}.}}
\\
\hline
\end{tabular}}

\caption{Case Study. Text in \textbf{\textcolor[rgb]{0.7,0.3,0.3}{red}} indicates errors and their causes in the response, while text in \textbf{\textcolor[rgb]{0.13,0.67,0.8}{blue}} marks key errors identified by ORION and the corresponding correct solutions.}
\label{case-study:casestudy-refinement}
\end{table*}
\begin{table*}[t]
\centering
\small
\renewcommand\arraystretch{1.2}
\setlength{\fboxsep}{1.0pt}
\resizebox{\linewidth}{!}{ 
\begin{tabular}{p{0.88\textwidth}}
\hline
\rowcolor{gray!8}{\textbf{Case 3:} Comparison of Responses from Different Baseline Methods.}\\ \hline
\textbf{Question:} Please reason step by step, and put your final answer within boxed\{\}. If \(m\) and \(n\) are integers such that \(3m + 4n = 100\), what is the smallest possible value of \(\lvert m - n \rvert?\) \\
\textbf{Ground-Truth:} 3 \\
\textbf{Vanilla Response:} To find the smallest possible value of \(\lvert m - n \rvert\) given \(3m + 4n = 100\), we solve for \(m\) in terms of \(n\):  
\(m = \frac{100 - 4n}{3}\). For \(m\) to be an integer, \(100 - 4n\) must be divisible by 3. Since \(100 \equiv 1 \pmod{3}\) and \(-4n \equiv -n \pmod{3}\), \textbf{\textcolor[rgb]{0.7,0.3,0.3}{we require \(n \equiv 0 \pmod{3}\).}} Let \(n = 3k\). Substituting gives  
\(m = \frac{100 - 12k}{3} = 33 - 4k\), so both \(m\) and \(n\) are integers: \(m = 33 - 4k, n = 3k\). Then  
\(\lvert m - n \rvert = \lvert (33 - 4k) - 3k \rvert = \lvert 33 - 7k \rvert\). Minimizing \(\lvert 33 - 7k \rvert\) reduces to choosing \(k\) nearest to \(33/7 \approx 4.714\). For \(k = 4\), \((m, n) = (17, 12)\) and \(\lvert m - n \rvert = 5\). For \(k = 5\), \((m, n) = (13, 15)\) and \(\lvert m - n \rvert = 2\). \textbf{\textcolor[rgb]{0.7,0.3,0.3}{Hence, the smallest possible value is boxed \{2\}.}}\\

\textbf{ORION w/o Self-Reflection:} 
Hmm, okay, let’s try to express \(n\) in terms of \(m\): \(n=\frac{100-3m}{4}\). For \(n\) to be an integer, \(100-3m\) must be divisible by 4. Since \(100\equiv 0 \pmod{4}\), we need \(3m\equiv 0 \pmod{4}\), and because 3 and 4 are coprime, \(m\) must be a multiple of 4. Let’s set \(m=4k\), giving \(n=25-3k\). Then \(|m-n|=|4k-(25-3k)|=|7k-25|\)... To minimize this, \(k\) should be near \(25/7\approx3.57\). Checking \(k=3\) gives \((m,n)=(12,16)\) and \(|m-n|=4\); checking \(k=4\) gives \((m,n)=(16,13)\) and \(|m-n|=3\). \textbf{\textcolor[rgb]{0.7,0.3,0.3}{Both are small, but the smaller one seems to be 4 when \(k=3\)... Thus, the final answer is boxed\{4\}.}}\\
\textbf{ORION w/o Solution Error:} 
\#\# Error Analysis: The main mistake in the original solution was incorrectly simplifying the equation \(3m + 4n = 100\) when attempting to express \(n\) in terms of \(m\), leading to non-integer values for \(m\) and \(n\).\\
\#\# Correct Reasoning: We want to minimize \(\lvert m - n \rvert\) given \(3m + 4n = 100\). Solving for \(m\),  
\(m = \frac{100 - 4n}{3}\). Suppose \(n\) is chosen to make \(m\) an integer. Trying \(n = 3k\), we get  
\(m = \frac{100 - 12k}{3} = 33 - 4k\). Approximating, we take \(m \approx 33 - 4k\). Then  
\(\lvert m - n \rvert \approx \lvert 33 - 7k \rvert\). \textbf{\textcolor[rgb]{0.7,0.3,0.3}{If \(k = 5\), this gives \(m = 13, n = 15\), so \(\lvert m - n \rvert = 2\). Thus, the smallest possible value is boxed \{2\}.}}\\
\textbf{ORION:}  \#\# Error Analysis: The possible mistake was the incorrect handling of \(3m + 4n = 100\) when solving for \(m, n\). The substitution and simplification produced a wrong expression for \(n\), leading to non-integer values. The error arose \textbf{\textcolor[rgb]{0.13,0.67,0.8}{because the expression for \(n\) was not properly simplified or checked for integer solutions.}}\\
\#\# Correct Reasoning: We aim to minimize \(\lvert m - n \rvert\) given \(3m + 4n = 100\). Expressing \(n\) in terms of \(m\),  
\(n = \frac{100 - 3m}{4}\). For \(n\) to be integer, \(100 - 3m\) must be divisible by 4. Since \(100 \equiv 0 \pmod{4}\), this requires \(3m \equiv 0 \pmod{4}\), so \(m = 4k\). Substituting gives  
\(n = 25 - 3k, \quad m = 4k\). For non-negative integers, \(0 \leq k \leq 8\). Then  
\(\lvert m - n \rvert = \lvert 4k - (25 - 3k) \rvert = \lvert 7k - 25 \rvert\). Evaluating: \(k = 3 \Rightarrow \lvert 21 - 25 \rvert = 4\), \(k = 4 \Rightarrow \lvert 28 - 25 \rvert = 3\), others yield larger values. \textbf{\textcolor[rgb]{0.13,0.67,0.8}{Thus the minimum is boxed \{3\}.}}\\
\hline
\end{tabular}}
\caption{Case Study. Text in \textbf{\textcolor[rgb]{0.7,0.3,0.3}{red}} indicates errors and their causes in the response, while text in \textbf{\textcolor[rgb]{0.13,0.67,0.8}{blue}} marks key errors identified by ORION and the corresponding correct solutions.}
\label{case-study:casestudy-different-method}
\end{table*}

\end{document}